\documentclass[10pt,twocolumn,letterpaper]{article}

\usepackage{wacv}
\usepackage{times}
\usepackage{epsfig}
\usepackage{graphicx}
\usepackage{amsmath}
\usepackage{amssymb}
\usepackage{subfig}
\usepackage{multirow,bigstrut}
\usepackage{amsmath}

\wacvfinalcopy 

\def\wacvPaperID{37} 

\ifwacvfinal\fi
\begin{document}

\title{Enriched Deep Recurrent Visual Attention Model for Multiple Object Recognition}

\author{Artsiom Ablavatski \hspace{2cm} Shijian Lu\\
Visual Attention Lab, Institute for Infocomm Research (I$^{2}$R), Singapore\\
{\tt\small stuaa@i2r.a-star.edu.sg, slu@i2r.a-star.edu.sg}
\and
Jianfei Cai \\
School of Computer Engineering, Nanyang Technological University, Singapore\\
{\tt\small asjfcai@ntu.edu.sg}
}

\maketitle
\ifwacvfinal\thispagestyle{empty}\fi

\begin{abstract}
We design an Enriched Deep Recurrent Visual Attention Model (EDRAM) --- an improved attention-based architecture for multiple object recognition. The proposed model is a fully differentiable unit that can be optimized end-to-end by using Stochastic Gradient Descent (SGD). The Spatial Transformer (ST) was employed as visual attention mechanism which allows to learn the geometric transformation of objects within images. With the combination of the Spatial Transformer and the powerful recurrent architecture, the proposed EDRAM can localize and recognize objects  simultaneously. EDRAM has been evaluated on two publicly available datasets including MNIST Cluttered (with 70K cluttered digits) and SVHN (with up to 250k real world images of house numbers). Experiments show that it obtains superior performance as compared with the state-of-the-art models.
\end{abstract}

\section{Introduction}
\let\thefootnote\relax\footnote{Publiished as a conference paper at 2017 IEEE Winter Conference on Applications of Computer Vision (WACV)}
Recurrent models of visual attention have demonstrated superior performance on a variety of recognition and classification tasks ~\cite{mnih2014recurrent,ba2014multiple,sonderby2015recurrent,xu2015show} in recent year. A recurrent model of visual attention is a task-driven agent interacting with a visual environment which observes the environment via a bandwidth-limited sensor at each time stamp. Recurrent models consist of  two crucial components: an attention mechanism and a recurrent network. The first component, a simple attention mechanism as introduced by Mnih \etal~\cite{mnih2014recurrent}, has demonstrated great success on recognizing digits within the Street View House Number (SVHN) dataset~\cite{goodfellow2013multi}. However, this mechanism is restricted by its simplicity which extracts a fixed amount of patches on each iteration with predefined scales.  The recently introduced more flexible and sophisticated visual attention mechanisms \cite{gregor2015draw,jaderberg2015spatial} achieved state-of-the-art results on the $60 \times 60$ MNIST Cluttered dataset \cite{sonderby2015recurrent}. The attention mechanisms advance in the patch extraction allowing to cover any 2D affine transformation of objects in the image. Hence the mechanisms are robust to most geometric transformations and have demonstrated superior performance comparable with the humans. The second component, the recurrent network also plays an essential in the recurrent models of visual attention. More powerful recurrent architecture employed in the Deep Recurrent Visual Attention Model \cite{ba2014multiple} significantly outperformed the original Recurrent Attention Model proposed by Mnih \etal~\cite{mnih2014recurrent} while leaving the rest of the systems identical.

\begin{figure*}
\centering
\begin{tabular}{ccc}
\multirow{2}{*}[1.01in]{\includegraphics[width=0.465\linewidth]{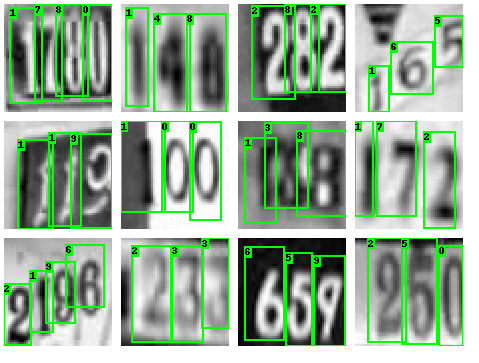}}&\includegraphics[width=0.16\linewidth]{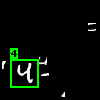}&\includegraphics[width=0.16\linewidth]{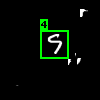} \\
&\includegraphics[width=0.16\linewidth]{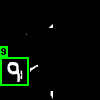}&\includegraphics[width=0.16\linewidth]{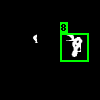} \\ \\
(a)&\multicolumn{2}{c}{(b)}\\
\end{tabular}
\caption{Examples of recognized images of the Street View House Numbers dataset~\cite{goodfellow2013multi} in (a) and the MNIST Cluttered dataset~\cite{mnih2014recurrent} in (b) where the green-color boxes show the localization and the digits at the box top-left corner show the recognition results.}
\label{fig:dataset_examples}
\end{figure*}

Inspired by deep recurrent visual attention model and the power of the visual attention mechanism \cite{jaderberg2015spatial}, we propose an Enriched Deep Recurrent Visual Attention model (EDRAM) that consists of a flexible and powerful attention mechanism along with a smart and light-weight recurrent neural network. The proposed technique is fully differentiable and has been trained end-to-end by using Stochastic Gradient Descent (SGD). It obtained superior performance as evaluated on two publicly available datasets including the multi-digit SVHN dataset \cite{goodfellow2013multi} as illustrated in Fig.~\ref{fig:dataset_examples}a and the MNIST Cluttered~\cite{mnih2014recurrent} as illustrated in Fig.~\ref{fig:dataset_examples}b.

\section{Related work}

Recurrent models of visual attention have been attracting increasing interest in recent years. The Recurrent Attention Model (RAM) proposed by Mnih \etal~\cite{mnih2014recurrent} employs a recurrent neural network (RNN) to integrate visual information (image patch or glimpse) over time. By REINFORCE optimization of the network~\cite{williams1992simple}, they achieved a huge reduction of computational cost as well as state-of-the-art performance on the MNIST Cluttered dataset. Ba \etal~\cite{ba2014multiple} extended the glimpse network for multiple object recognition with visual attention. They introduced the Deep Recurrent Visual Attention Model (DRAM) that integrates a simple visual attention mechanism with the neural network based on Long Short-Term Memory (LSTM) gated recurrent unit~\cite{hochreiter1997long}. The REINFORCE learning rule, employed in~\cite{mnih2014recurrent} to train their attention model, was used to learn the network ``{\em where}'' and ``{\em what}''. Though the DRAM achieved superior result in a number of tasks such as the MNIST pair classification and SVHN recognition, the attention mechanism used is straightforward by extracting patches of fixed scales only, where the potential of the visual attention mechanism is far from fully exploited.

In comparison to the simple attention mechanism, Gregor \etal~\cite{gregor2015draw} introduced Selective Attention Model that created a differentiable, end-to-end trainable system which greatly improves the recognition accuracy on the MNIST Cluttered dataset. The idea of this attention mechanism is to position $N \times N $ set of Gaussian filters forming a grid centered at the particular spatial coordinates with which rectangle patches of different scales can be extracted. More recently, Jaderberg \etal~\cite{jaderberg2015spatial} proposed the Spatial Transformer Network that can deal with not only scales, but also any 2D affine transformation of objects in images. Using Spatial Transformer in combination with the convolutional neural network, Jaderberg \etal~\cite{jaderberg2014synthetic} achieved state-of-the-art result on the SVHN dataset -- 3.6\% error rate.

Though different deep attention models have been designed, all existing models have various constraints. For example, the Differentiable RAM~\cite{gregor2015draw} is fully differentiable and can be trained using SGD, but the model has a weak network architecture and does not scale well to real world tasks. The DRAM~\cite{ba2014multiple} has a powerful architecture, but the sampling strategy makes the whole network non-differentiable. Building on the long line of the previous attempts of attention-based visual processing methods~\cite{ba2014multiple,mnih2014recurrent,jaderberg2015spatial,sonderby2015recurrent,gregor2015draw}, the proposed EDRAM expands the idea of using recurrent connections inside the attention mechanism~\cite{sonderby2015recurrent} and improves the Deep Recurrent Visual Attention Model (DRAM)~\cite{ba2014multiple} from several aspects as follows:
\begin{itemize}  
\item It made previously non-differentiable architecture fully differentiable by using the Spatial Transformer. It allowed us to optimize the network parameters end-to-end by using SGD with backpropagation framework.
\item A flexible loss function was designed based on Cross Entropy and Mean Squared Error function with which the EDRAM can be trained efficiently.
\item It obtains superior performance on the MNIST Cluttered and SVHN datasets as compared with state-of-the-art methods.
\end{itemize}

\section{Enriched Deep Recurrent Visual Attention Model}

The underlying idea of the EDRAM is to combine a powerful and computationally simple recurrent neural network with a flexible and adjustable attention mechanism (ST), while making the whole network fully differentiable and trainable through SGD.

\subsection{Network Architecture}

\begin{figure}
\begin{center}
\includegraphics[width=0.8\linewidth]{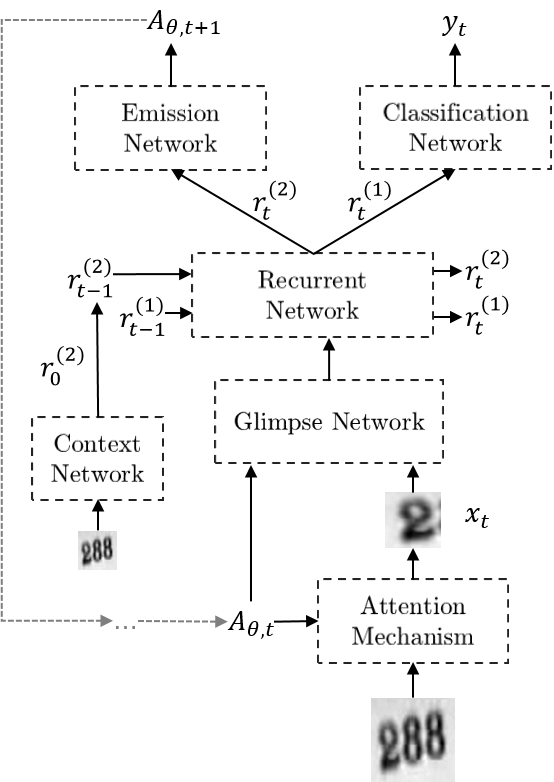}
\end{center}
\caption{The architecture of the EDRAM.}
\label{fig:edram_scheme}
\end{figure}

Inspired by the model proposed by Ba \etal~\cite{ba2014multiple}, our network architecture was designed to satisfy the complexity requirements and the capability to learn a very nonlinear classification function. It can be decomposed into several sub-components including an attention mechanism, a context network, a glimpse network, a recurrent network, a classification network and an emission network as illustrated in Fig.~\ref{fig:edram_scheme}. Each sub-component can be referred by a ``network'' because it is a typical a multi-layered neural network.

The context network receives a down-sampled low-resolution image as input and processes it through a three-layered convolutional neural network. It produces a feature vector $r_0^{(2)}$ that serves as an initialization of a hidden state of the second LSTM unit in the recurrent network. 

The attention mechanism reads an image patch $x_t$ by using the transformation parameters $A_{\theta,t}$ (more details to be described in Section 3.2) that have been predicted on the previous iteration. Parameters for the first iteration are defined in a way that the attention mechanism reads the whole image without transformation. An algorithm for the read operation and transformation parameters will be defined in the next subsection.

Different from the DRAM where the glimpse network is responsible only for producing discriminative features for the classification network, the glimpse network in EDRAM integrates the localization network from the Spatial Transformer. Therefore, the glimpse network in the EDRAM  contains a number of convolution layers followed by the max pooling layer. The number of convolutions varies and depends on the difficulty of a recognition task. The result of the convolution layers and the transformation parameters $A_{\theta,t}$ are used in multiple isolated fully-connected layers. The outputs of the isolated fully-connected layers are combined together by element-wise multiplication to form the final glimpse feature vector. This type of combination ``what'' and ``where'' was proposed by Larochelle and Hinton \etal~\cite{larochelle2010learning}.

The recurrent network contains two LSTM units stacked one above the other with hidden states $r_t^{(1)}$ and $r_t^{(2)}$. The first LSTM receives the glimpse feature vector and uses the hidden state $r_t^{(1)}$ to produce a feature vector for the classification network. Based on the hidden state of the first LSTM $r_t^{(1)}$ and the hidden state $r_t^{(2)}$, the second LSTM unit produces a feature vector for the emission network. The hidden state $r_t^{(1)}$ of the first LSTM is independent of the hidden state $r_t^{(2)}$ of the second LSTM and the glimpse parameters $A_{\theta,t}$. This means that a prediction of the classification network depends only on an extracted patch and is independent of a location and the transformation parameters.

The classification and emission networks map feature vectors from different layers of the recurrent network (by using fully-connected layers) to predict labels $y_t$ and the transformation parameters $A_{\theta,t+1}$ for the next iteration, respectively. The classification network has two fully-connected layers followed by softmax output layer. The emission network employs the fully-connected layer to predict the transformation parameters.

The EDRAM processes the image in a sequential manner with $T$ steps. At each time step $t$, the model receives the parameters of transformation $A_{\theta,t}$ and uses the attention mechanism to extract the patch $x_t$ at the location as defined by the parameters $\theta_3,\theta_6$  (to be described in Section 3.2)  in the transformation matrix $A_{\theta,t}$. The model uses the observation $x_t$, processed by the glimpse network, and the parameters $A_{\theta,t}$ to update its internal states $r_t^{(1)}, r_t^{(2)}$ and produce the parameters $A_{\theta,t+1}$ for the next step. Besides that, the model makes a prediction $y_t$ based on the internal states $r_t^{(1)}$ of the first LSTM. The attention mechanism controls the number of pixels in the patch $x_t$, which is usually much smaller than the number of pixels in the original image.

\subsection{Spatial Transformer attention mechanism}

Instead of using the nondifferentiable attention mechanism that simply reads patches at the given location ${(x,y)}$, our Spatial Transformer attention mechanism is inspired by the differentiable Spatial Transformer Networks as proposed by Jaderberg \etal~\cite{jaderberg2015spatial}. The original Spatial Transformer Networks only contains a localization network, grid generator and sampler. All parts are fully differentiable that enables the optimization of the whole model end-to-end by using gradient descent within a standard backpropagation framework.
\begin{figure}
\begin{center}
\includegraphics[width=0.8\linewidth]{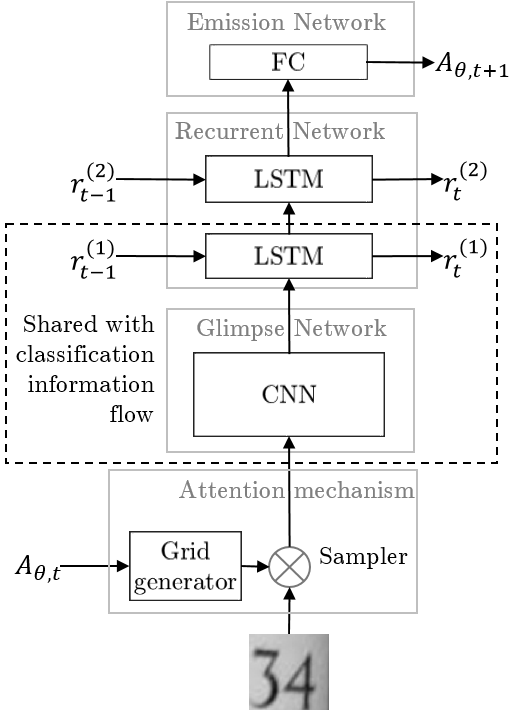}
\end{center}
\caption{The localization network of the EDRAM with shared layers: CNN - convolutional layers, followed max pooling layer; LSTM - Long Short-Term Memory;  FC - Fully-connected layer.}
\label{fig:stam}
\end{figure}

The major constraint of the original Spatial Transformer Networks is that the transformation parameters for the grid generator were obtained using the standalone localization network, which introduces a larger number of parameters and increases the computational cost. Another drawback of the Spatial Transformer Networks is the supervision for the prediction when it is used in an iterative manner: the Spatial Transformer Networks was introduced only for feed-forward networks and predicts the transformation parameters only based on the current input data without using the information from the previous steps. The transformation parameters with a better accuracy could be obtained when the Spatial Transformer can incorporate information from the previous steps.

To overcome the downsides of the original Spatial Transformer Networks we propose a solution that incorporates information from previous iterations and at the same time reduces the number of parameters needed for the localization network. Our Spatial Transformer attention mechanism integrates the localization network into the glimpse network, which is responsible for both classification and transformation information flow. Technically, it means that backpropagation of the classification error through the glimpse network will affect on the localization error of the emission network and backpropagation of the localization error will affect on the prediction accuracy of the classification network. A flexible loss function therefore needs to be designed for the EDRAM to control the backpropagation of the errors and to learn both accurate transformation parameters and correct classification labels. Besides, sharing network parameters between the prediction and transformation information flows results in the reduction of the computational cost. The overall scheme of the localization network of the EDRAM is illustrated in Fig.~\ref{fig:stam}     

The Spatial Transformer is responsible for an affine transformation (zoom, rotation and skew) of mesh grid points $x^g_i,y^g_i$ according to the parameters  $A_{\theta,t}$.
\begin{equation}
  A_{\theta,t} = \begin{bmatrix} \;
 					\theta_{1,t} & \theta_{2,t} & \theta_{3,t} \; \\
 					\theta_{4,t} & \theta_{5,t} & \theta_{6,t} \; \\
  					\end{bmatrix} \;
\end{equation}
\begin{equation}
\begin{pmatrix} x^s_i \\ y^s_i \end{pmatrix} = A_{\theta,t}\begin{pmatrix} x^g_i \\ y^g_i \\ 1 \end{pmatrix}= \begin{bmatrix} \theta_{1,t} \theta_{2,t} \theta_{3,t} \\ \theta_{4,t} \theta_{5,t} \theta_{6,t} \end{bmatrix} \begin{pmatrix} x^g_i \\ y^g_i \\ 1 \end{pmatrix}
\end{equation}
where $\theta_{1,t},\theta_{5,t}$ -- determine a zoom, $\theta_{2,t},\theta_{4,t}$ -- determine a skewness in $x,y$ directions respectively and $\theta_{3,t},\theta_{6,t}$ -- determine the center position of the mesh grid.

Since the mesh grid of points (Grid generator in Fig.~\ref{fig:stam}) does not correspond exactly to one particular point $x^s_i,y^s_i$ in the input image, the bilinear interpolation  (Sampler in Fig.~\ref{fig:stam}) is used to output the fixed scale patch $X_t$ from the input image $I$ for further processing.
\begin{equation}
X_t=\sum_{n}^{H}\sum_{m}^{W}I_{nm}\text{max}(0,1-\lvert x^s_i-m \rvert)\text{max}(0,1-\lvert y^s_i-n \rvert)
\label{eq:2} 
\end{equation}
Then the partial derivatives for the bilinear sampling \eqref{eq:2} w.r.t the sampling grid coordinates can be formulated as follows:
\begin{equation}
\frac{dX_t}{dx^s_i}=\sum_{n}^{H}\sum_{m}^{W}I_{nm}\text{max}(0,1-\lvert y^s_i-n \rvert)\begin{cases}
      0&\text{if}\ \lvert m-x^s_i \rvert \geq 1 \\
      1&\text{if}\ m \geq x^s_i \\
      -1&\text{if}\  m<x^s_i
\end{cases} 
\end{equation}
\begin{equation}
\frac{dX_t}{dy^s_i}=\sum_{n}^{H}\sum_{m}^{W}I_{nm}\text{max}(0,1-\lvert x^s_i-m \rvert)\begin{cases}
      0&\text{if}\ \lvert n-y^s_i \rvert \geq 1 \\
      1&\text{if}\ n \geq y^s_i \\
      -1&\text{if}\  n<y^s_i
\end{cases} 
\end{equation}
The sub-differentiable sampling allows backpropagation of the loss to the sampling grid coordinates which leads to flow back the gradients to the transformation parameters $A_{\theta,t}$ and to Emission Network in Fig.~\ref{fig:stam}. In addition, to encourage the Spatial Transformer attention mechanism to learn more accurate transformation parameters we allow backpropagation of the loss in opposite direction through the supervision of the parameters obtained on the previous iteration (see Learning ``Where" in the next section). This contribute in a precise localization of the extracted patches by the attention mechanism after only a few epochs of training.

The Spatial Transformer attention mechanism is fully differentiable and satisfies both requirements of flexibility and adjustability. Hence, it allows us to train the attention mechanism  with standard backpropagation.

\subsection{Learning ``Where'' and ``What''}

In the context of multiple object recognition, the network should locate the necessary objects of an image (``Where'') and successfully recognize them (``What'') in order to achieve the desired performance. Hence, the objective function should penalize any false positives predictions as well as incorrect recognitions at true positive locations. 

The loss function is designed to force the EDRAM to recognize necessary objects in a finite number of steps. For each object in the image we allow the network to make a fixed number of predictions $N$. The network produces a final class prediction for the given object by averaging the predictions. Suppose we have $S$ targets in the image, the loss function will be calculated only for $N \; \times \; S \;$ steps:
\begin{equation}
L =  \frac{1}{N}\displaystyle\sum_{i=1}^{S} \displaystyle\sum_{j=1}^{N} L_{i,j} \; \label{eq:1} 
\end{equation}
The loss function $L_{i,j}$ for each iteration includes a weighted summation of the Cross Entropy loss for the given glimpse  $L_{i,j}^{y}$ and weighted Mean Squared Error of the transformation parameters $L_{i,j}^{A_{\theta}}$:
\begin{equation}
L_{i,j} = \alpha_1*L_{i,j}^{y} + \alpha_2*L_{i,j}^{A_{\theta}} \;
\end{equation}
where $L_{i,j}^{y}$ can be interpreted as ``{\em what to look}'' and $L_{i,j}^{A_{\theta}}$ --- ``{\em where to look}''. The hyperparameters $\alpha_1$ and $\alpha_2$ give a good trade-off between classification and transformation loss, forcing the model to simultaneously optimize for a better patch extraction and for better recognition.
\begin{equation}
L_{i,j}^{y} = -\log {p_{i,j,y_{gt}} }\;
\end{equation}
\begin{equation}
L_{i,j}^{A_{\theta}} = \displaystyle\sum_{k=1}^{6}\beta_k*(\theta_{k,i,j} - \theta_{k,i,j}^{gt})^2 \;
\end{equation}
where $p_{i,j,y_{gt}}$ is a predicted class probability on a ground-truth position,  $\theta_{1,i,j},\dots,\theta_{6,i,j}$ --- elements of the matrix $A_{\theta}$ and $\theta_{1,gi,j}^{gt},\dots,\theta_{6,i,j}^{gt}$ --- ground truth values for iteration $i,j$. The hyperparameters $\beta_1,\dots,\beta_6$ force the network to pay more attention to critical parameters such as width, height and coordinates of the mesh grid and ignore unimportant parameters such as the skewness.

\section{Experiments}

EDRAM has been evaluated over the MNIST Cluttered dataset~\cite{mnih2014recurrent,ba2014multiple} as well as a real-world object recognition task by using the Street View House Numbers (SVHN) dataset~\cite{netzer2011reading}.

Since the attention mechanism is fully differentiable, the proposed network is trained with standard backpropagation and SGD by using Adam optimization algorithm~\cite{kingma2014adam}. Gradient step clipping techniques are applied~\cite{mikolov2012statistical,pascanu2012difficulty} to ensure the absence of the gradient exploding during the learning over the recursive structures. The value of the thresholded norm is chosen equal to 10. Following Cooijmans ~\cite{cooijmans2016recurrent}, the Batch Normalization proposed by Iofee and Szegedy~\cite{ioffe2015batch} is used on the MNIST Cluttered dataset to estimate the statistics independently for each iteration. Theano~\cite{bastien2012theano}, Blocks and Fuel~\cite{van2015blocks} are used to implement and conduct the experiments with the MNIST Cluttered and SVHN datasets.

For comparison, we take the results from the latest works by Almahairi \etal~\cite{almahairi2015dynamic} --- Dynamic Capasity Networks (DCN), Jaderberg \etal~\cite{jaderberg2015spatial} --- ST-CNN, Ba \etal~\cite{ba2014multiple} --- DRAM and the first result obtained on the SVHN dataset by Goodfellow \etal~\cite{goodfellow2013multi} --- 11 layer CNN. {\bf DCN} is an attention-based ensemble of convolutional networks of different capacities. The family of DRAM models includes the original Deep Recurrent Visual Attention Model ({\bf Single DRAM}), DRAM with Monte Carlo sampling policy ({\bf Single DRAM MC avg.}) and ensemble of two models of different recognition order ({\bf forward-backward DRAM MC avg.}). The {\bf 11 layer CNN} and ST-CNN are the feed-forward networks containing several convolutions layers followed by fully-connected layers. In addition, the ST-CNN includes one ({\bf ST-CNN Single}) or several ({\bf ST-CNN Multi}) Spatial Transformers between convolutional layers with separate localization networks. 

\subsection{MNIST Cluttered}

\begin{figure}
\begin{center}
\includegraphics[width=0.8\linewidth]{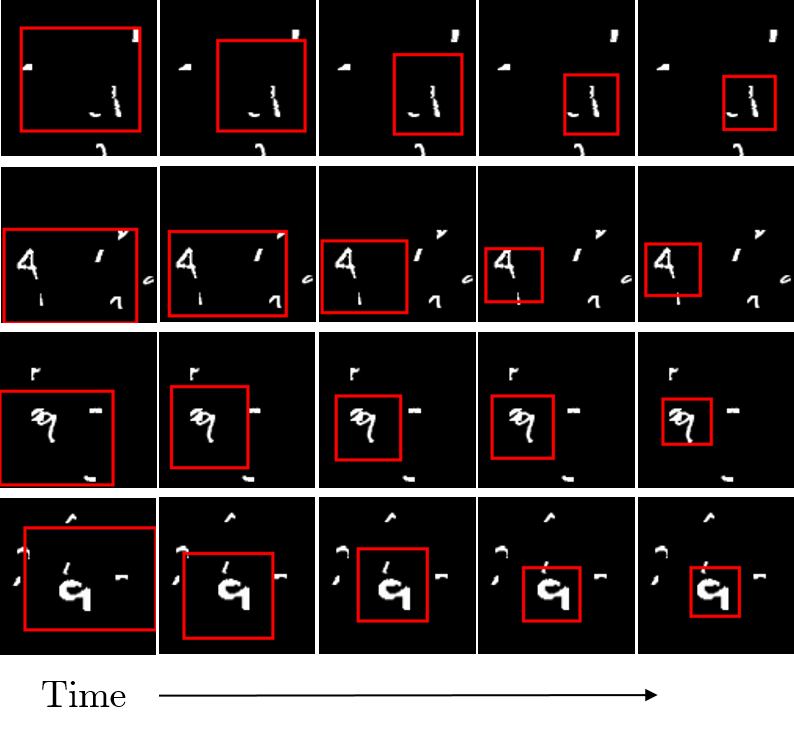}
\end{center}
\caption{MNIST Cluttered classification: Each row shows a sequence of glimpses taken by the network while recognizing MNIST Cluttered dataset. The red rectangle illustrates the location and size of the image patch extracted by the attention mechanism.}
\label{fig:mnist_attention}
\end{figure}

We first evaluate EDRAM on the $100 \times 100 \;$ MNIST Cluttered dataset~\cite{mnih2014recurrent}, where each image contains randomly located hand-written digit surrounded with digit-like fragments. The dataset has 60000 images for training and 10000 for testing.

At each time step, a glimpse of the size  $26 \times 26 \;$ from the input image is fed to the network and the model predicts parameters of extraction for the next iteration as well as a class label. The model produces the final classification result after a fixed number of glimpses (6 in our case).
In this experiment, we use ReLU activations for all layers except the recurrent network, where standard $tanh$ activation in LSTM units are employed. Context network takes down-sampled $12 \times 12 \;$ image and projects it into a vector using 3 convolutional layers (without any activations) with filter sizes $\left\{ 5,3,3 \right\}$ and number of filters $\left\{ 16,16,32 \right\}$, respectively. 
  
We use 6 convolutional layers in the glimpse network. The size of filters in each convolution of the glimpse network is chosen to be 3 and the numbers of filters in 6 convolutions are 64, 64, 128, 128, 160 and 192.
The max pooling is made with size $\left\{ 2,2 \right\}$ and stride  $\left\{ 2,2 \right\}$ after second and fourth convolutional layers. Zero-padding of half of the filter size is used in the first, third, forth and fifth convolutions of the glimpse network. There are 512 LSTM units and 1024 hidden units in each fully-connected layer of the model. 

Optimal values for hyperparameters $\alpha_1$ and $\alpha_2$ were found to be 1 by random search. To encourage the network to learn precise location of the target the weights $\beta_1,\beta_3,\beta_5$ and $\beta_6$ set to 1 whereas $\beta_2$ and $\beta_4$ are 0.5. A learning rate of $10^{-4}$ is used for training the model and exponentially reduced by a factor of 10 when the training loss
plateaus. The model is initialized with a uniform distribution for recurrent and convolutional units with the range of $[-0.01, 0.01] \;$ and a Gaussian distribution for fully-connected layers with a variance $0.001$. A mini-batch size of $128$ is used to estimate the gradient directions.
\begin{table}[htb]
\begin{center}
\begin{tabular}{lc}
\hline
Model & Test Error\\
\hline\hline
Convolutional, 2 layers & 14.35\%\\
RAM~\cite{mnih2014recurrent}, 8 glimpses, $12 \times 12 \;$, 4 scales & 8.11\%\\
Differentiable RAM~\cite{gregor2015draw}, 8 glimpses, $12 \times 12 \;$ & 3.36\%\\
ST-CNN Single~\cite{jaderberg2015spatial} & 1.7\% \\
DCN~\cite{almahairi2015dynamic}, 8 glimpses, $14 \times 14 \;$ & 1.39\%\\
\hline
Ours (6 glimpses,  $26 \times 26 \;$) & {\bf 0.6}\%\\
\end{tabular}
\end{center}
\caption{Recognition results on MNIST Cluttered dataset.}
\label{table:mnist_cluttered}
\end{table}

The results in Table~\ref{table:mnist_cluttered} demonstrate more then 2$\times$ improvement in the test error as compared to the state-of-the-art models on the MNIST Cluttered dataset. With the help of the proposed Spatial Transformer attention mechanism and the designed objective function, the network is able to learn where to find a digit in the cluttered background and how to recognize it accurately. Moreover, making the network fully differentiable allows it, to train end-to-end by standard back propagation and to use the batch normalization, which leads to faster convergence.  Fig.~\ref{fig:mnist_attention} illustrates the process of the attention mechanism where each row shows how the Spatial Transformer attention mechanism locates a digit on the image iteration by iteration accurately.
\begin{figure}[h]
\begin{center}
%
\includegraphics[width=0.8\linewidth]{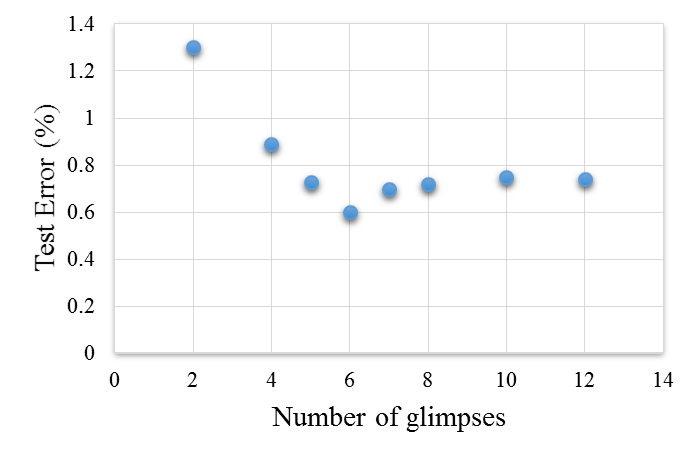}
\end{center}
\caption{The dependency of the number of processed glimpses vs accuracy on the MNIST Cluttered test dataset.}
\label{fig:glimpses}
\end{figure}

We show in Fig.~\ref{fig:glimpses} how the test error on the MNIST Cluttered dataset decreases when the number of glimpses is increased. We can see that the test error is decreasing almost linearly with increasing the number of patches and after 6 glimpses it saturates and performance does not improve significantly. So, 6 glimpses is a good trade-off between the accuracy and computation cost of the model.

\subsection{SVHN}

We also evaluate EDRAM on the multi-digit SVHN dataset and compared it with the state-of-the-art models. The SVHN dataset contains around $250k \;$ real world images of digits taken from pictures of house fronts. There are between 1 and 5 digits in each image, with a large variability in scale and spatial arrangement. Following the experimental setup in~\cite{goodfellow2013multi,ba2014multiple}, the test set is formed from $13k \;$ images and the rest of data (train and extra sets) is used to train the networks.

The data is preprocessed by generating tightly cropped $64 \times 64 \;$images with multi-digits at the center and similar data augmentation is used to create $54 \times 54 \;$jittered images during training. Similar to \cite{ba2014multiple}, RGB images are converted into grayscale. The model is trained to classify all the digits in an image sequentially with objective function as defined in Eq.~\eqref{eq:1}. Patches are given to classify each digit in the image. As images in the SVHN dataset have at most $5$ digit sequences, the overall amount of iterations is $18$ that equals $5 \times 3 \;$ plus $3$ patches for a terminal label.

The heuristic of learning two separate models of different reading orders (forward, backward) as proposed in~\cite{ba2014multiple} is adapted. As the localization network is integrated in the glimpse network and the transformation parameters are different for the models, the weights are not shared between forward and backward models. Fig.~\ref{fig:dataset_examples}a illustrates how the proposed EDRAM extracts glimpses accurately around digits.

In the experiments, a square extraction window of size $34 \times 34 \;$ pixels is used. Initialization of the network is chosen identically to the MNIST Cluttered experiment. An initial value for learning rate is $10^{-4}$. A mini-batch size of $128$ is used to estimate the gradient direction. For all units except LSTM blocks, ReLU activation function was used. For the stacked LSTM blocks, standard $tanh$ activation was used. As SVHN dataset provides only size and location for digit bounding box, only 4 parameters $\theta_1,\theta_3,\theta_5,\theta_6$ from transformation matrix $A_{\theta}$ are used to estimate loss function $L_{i,j}^{A_{\theta}}$. The parameters of skewness  $\theta_2$ and  $\theta_2$ in the proposed system are learned by themselves for better prediction accuracy. The loss function hyperparameters are chosen to be identical to the experiment with MNIST Cluttered dataset.
The training took 5 days on a single modern GPU.
\begin{table}[htb]
\begin{center}
\begin{tabular}{lc}
\hline
Model&Test Error \\
\hline
11 layer CNN~\cite{goodfellow2013multi}& 3.96\% \\
Single DRAM~\cite{ba2014multiple}& 5.1\% \\
Single DRAM MC avg.~\cite{ba2014multiple}& 4.4\% \\
forward-backward DRAM MC avg.~\cite{ba2014multiple}&3.9\% \\
ST-CNN Single~\cite{jaderberg2015spatial}& 3.7\% \\
ST-CNN Multi~\cite{jaderberg2015spatial}& {\bf 3.6}\% \\
\hline
Ours (Single model) & 4.36\% \\
Ours (forward-backward ensemble)& {\bf 3.6}\% \\
\end{tabular}
\end{center}
\caption{Sequence recognition error rates on SVHN dataset.}
\label{table:svhn}
\end{table}

The proposed approach obtained state-of-the-art performance in recognition of multiple objects from the real world as shown in Table~\ref{table:svhn} while having 1.7$\times$ less parameters ($37M/22M\approx 1.7$) than previous state-of-the-art model ST-CNN Multi (see Table~\ref{table:models_params}). This proves the effectiveness of the developed objective function that forces the network to locate the desired objects on images and successfully recognize them one by one. The usage of the attention mechanism allows the network to ignore redundant information from images and extract patches that are necessary for the prediction of the correct class labels.
\begin{table}[htb]
\begin{center}
\begin{tabular}{lcccc}
\hline
Model & Parameters (millions) \\
\hline
10 layer CNN & 51 \\
Single DRAM~\cite{ba2014multiple} & 14 \\
Single DRAM MC avg.~\cite{ba2014multiple} & 14 \\
forward-backward DRAM MC avg.~\cite{ba2014multiple} & 28 \\
ST-CNN Single~\cite{jaderberg2015spatial} & 33 \\
ST-CNN Multi~\cite{jaderberg2015spatial} & 37 \\
\hline
Ours (Single model) & {\bf 11}  \\
Ours (forward-backward ensemble) & {\bf 22} \\
\end{tabular}
\end{center}
\caption{Computation cost of different Deep Convolutional Networks.}
\label{table:models_params}
\end{table}

The computational cost of the neural networks (NN) depends on the number of parameters, with more parameters the model needs more space to be stored and more floating-point operations (FLOPs) to execute to produce the final output. This creates a difficulty of applying NNs on different embedded platforms with limited memory and processing units, like mobile phones ~\cite{han2015learning}. Besides that, significant redundancy has been reported in many state-of-the-art neural network models ~\cite{denil2013predicting}. The integration of the separate localization network into the glimpse network allows to achieve a significant computation cost reduction in comparison with other state-of-the-art models. Table~\ref{table:models_params} shows the number of parameters of the proposed model in comparison with other deep convolutional neural networks. Though we only matched the performance on the SVHN dataset, our network contains $1.7$ times less parameters than the state-of-the-art ST-CNN Multi.

\section{Conclusions}

This paper presents an Enriched Deep Recurrent Visual
Attention Model that is fully differentiatiable and trainable end-to-end using SGD. The EDRAM outperforms the state-of-the-art result on the MNIST Cluttered dataset and matches the state-of-the-art models on a multi-digit house number recognition task. It requires a smaller amount of parameters and less computation resources, thereby proving that attention mechanism has a big impact on accuracy and efficiency of the model.

{\small
\bibliographystyle{ieee}
\bibliography{egbib}
}

\end{document}